%% file: root.tex
\documentclass[letterpaper, 10 pt, conference]{ieeeconf}

\IEEEoverridecommandlockouts
\usepackage{afterpage}
\usepackage{xspace}
\usepackage{amssymb,paralist,epsfig,standalone,bm,placeins}  % assumes amsmath package installed
\usepackage{graphicx} % for pdf, bitmapped graphics files
\usepackage{xcolor}
\usepackage{multirow}
\usepackage{makecell}
\usepackage{multicol}
\usepackage[utf8]{inputenc}
\usepackage{amsmath,amsfonts,booktabs,cite} % general useful stuff
\usepackage{siunitx,textcomp} % for the degree symbol
\usepackage[hidelinks]{hyperref} % to have hyperlinks and for the \ref macros
\let\labelindent\relax
\usepackage[inline]{enumitem} % inline enumerate
\usepackage[nolist]{acronym} % acronyms by the \ac{label} command
\usepackage{caption}
\usepackage{tikz,standalone,pgfplots}
\usetikzlibrary{patterns}
\usepackage{pgfplots, pgfplotstable}
\graphicspath{{figures_tex/}}

\usepackage{changes}
\usepackage[ruled,vlined]{algorithm2e}

\usepackage{subcaption}
\usepackage{fp}
\usepackage{forloop}
\usepackage{siunitx} %for printing
\usepackage{contour}
\usepackage{color}
\usepackage{booktabs} % for professional tables
\usepackage{multirow} % For merging multiple rows
\usepackage{mathtools} %for multlined
\usepackage{footmisc}  % For footnote refs

\makeatletter
\newcommand{\removelatexerror}{\let\@latex@error\@gobble}
\makeatother
\pgfplotsset{compat=1.12}
\input{acronyms.tex}
\input{macro.tex}

\graphicspath{{img_tex/}}

\definecolor{findOptimalPartition}{HTML}{D7191C}
\definecolor{storeClusterComponent}{HTML}{FDAE61}
\definecolor{dbscan}{HTML}{ABDDA4}
\definecolor{constructCluster}{HTML}{2B83BA}
\title{\LARGE \bf
Evaluating the Pre-Dressing Step: Unfolding Medical Garments via Imitation Learning
}

\author{David Blanco-Mulero, Julia Borras, Carme Torras%
\thanks{This work was financially supported by the European Union's Horizon Europe programme project SoftEnable (grant agreement No. 101070600).}
\thanks{The authors are with the Institut de Robòtica i Informàtica Industrial, CSIC-UPC, 08028, Barcelona, Spain. (e-mail: \texttt{david.blanco.mulero@upc.edu})}
}

\begin{document}
\maketitle
\thispagestyle{empty}
\pagestyle{empty}

%%%%%%%%%%%%%%%%%%%%%%%%%%%%%%%%%%%%%%%%%%%%%%%%%%%%%%%%%%%%%%%%%%%%%%%%%%%%%%%%

\begin{abstract}
Robotic-assisted dressing has the potential to significantly aid both patients as well as healthcare personnel, reducing the workload and improving the efficiency in clinical settings.
While substantial progress has been made in robotic dressing assistance, prior works typically assume that garments are already unfolded and ready for use.
However, in medical applications gowns and aprons are often stored in a folded configuration, requiring an additional unfolding step.
In this paper, we introduce the \textit{pre-dressing} step, the process of unfolding garments prior to assisted dressing.
We leverage imitation learning for learning three manipulation primitives, including both high and low acceleration motions.
In addition, we employ a visual classifier to categorise the garment state as closed, partly opened, and fully opened.
We conduct an empirical evaluation of the learned manipulation primitives as well as their combinations.
Our results show that highly dynamic motions are not effective for unfolding freshly unpacked garments, where the combination of motions can efficiently enhance the opening configuration.

\end{abstract}

%%%%%%%% Start of document

\section{Introduction}\label{sec:introduction}
\input{sections/introduction}

\section{Related work}\label{sec:related_work}
\input{sections/related_works}

\section{Pre-Dressing Manipulation and Metrics}\label{sec:method}
\input{sections/method}

\begin{figure*}
\centering
\def\svgwidth{\linewidth}
{
    \vspace{2mm}
    \fontsize{9}{9}%\selectfont\sf
    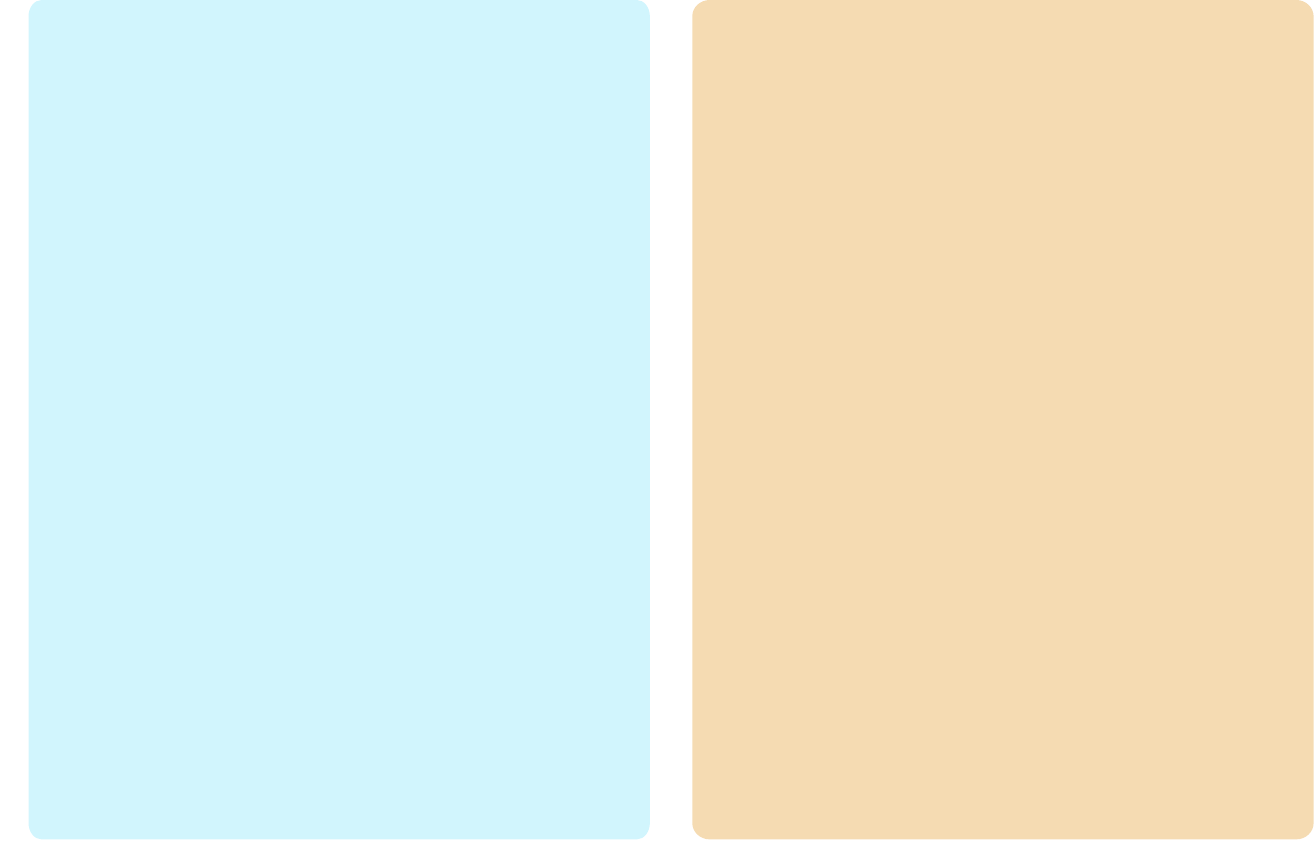}
    \caption{Qualitative results for the three manipulation primitives in (a) previously opened gowns and (b) unpacked gowns. The first row shows the gown before execution, and the second row after reaching an opened or a partly opened state.
    We indicate the category predicted by the visual classifier and its confidence, as well as the number of iterations it took to reach either an opened state, or, if no further improvement occurred, the first iteration at which a partial state was reached.
    \vspace{-2mm}
    } 
\label{fig:exp-primitives}
\end{figure*}
\section{Experiments}\label{sec:experiments}
\input{sections/experiment_setup}

\input{sections/results}

\section{Discussion}\label{sec:discussion}
\input{sections/discussion}

\section{Conclusions}
\label{sec:conclusions}
In this work, we introduced the task of pre-dressing, the  step of unfolding medical garments starting in a folded configuration before assisted dressing.
We evaluated three manipulation primitives for pre-dressing: fling, shake, and twist, learned from human demonstrations through the \ac{dmp} framework.
To assess the garment opening, we introduced three categories: closed, partly opened, and opened, and trained a visual classifier for state recognition.

Our experiments evaluated the effectiveness of the learned primitives in unfolding medical gowns, both individually, in combination with each other, as well as with a quasi-static motion.
Our results showed that individual primitives could effectively unfold previously opened gowns in few iterations, while they required more attempts to partly open unpacked gowns.
Here, the combination of different primitives proved more efficient to reach the partly opened state.
While our results showed that neither high- nor low-acceleration motions could break the electrostatic forces to unfold unpacked gowns, their combination was more effective for
reaching a sufficient partly opened state for assisted dressing.

Future work could explore combining the dynamic and quasi-static motions presented here with visual-tactile methods or re-grasping strategies to enhance the unfolding success.
Building on this, we aim to integrate the pre-dressing step into a robotic-assisted dressing pipeline, with the broader objective of supporting nurses in hospital settings.

\bibliographystyle{IEEEtran}
\bibliography{literature, refs}

\end{document}

%% file: acronyms.tex
\newacro{rl}[RL]{Reinforcement Learning}
\newacro{il}[IL]{Imitation Learning}
\newacro{dmp}[DMP]{Dynamic Movement Primitive}
\newacro{dof}[DoF]{Degree of Freedom}
\acrodefplural{dof}[DoFs]{Degrees of Freedom}
\newacro{pca}[PCA]{Principal Component Analysis}
\newacro{ik}[IK]{Inverse Kinematics}

%% file: macro.tex
\newcommand{\figref}[1]{\hyperref[#1]{Fig.~\ref*{#1}}}
\newcommand{\tabref}[1]{\hyperref[#1]{Table~\ref*{#1}}}
\newcommand{\secref}[1]{\hyperref[#1]{Section~\ref*{#1}}}
\newcommand{\algoref}[1]{\hyperref[#1]{Algorithm~\ref*{#1}}}

\newlength{\Oldarrayrulewidth}

%% file: sections/introduction.tex
Dressing assistance is a crucial task performed by caregivers to assist individuals in nursing facilities or hospitals.
Beyond patient care, dressing is also crucial in pre-surgery rooms and intensive care units, where nurses assist each other dress up in protective gowns.
Robotic-assisted dressing~\cite{zhang_2022_assisted_dressing_science} has emerged as a promising solution to address the shortage of healthcare personnel while also reducing the burden of repetitive tasks on caregivers and nurses.

\begin{figure}
\centering
\def\svgwidth{\linewidth}
{
    \fontsize{9}{9}%\selectfont\sf
    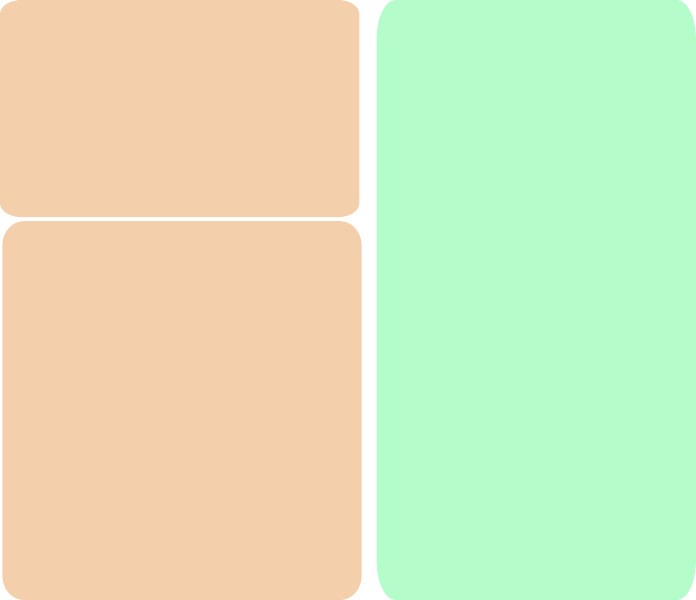}
    \caption{The \textit{pre-dressing} step is the transition from an initially folded garment to an unfolded garment prepared for assisted dressing. Here, we utilise imitation learning for learning manipulation primitives that perform the garment unfolding while adhering to the robot hardware limits.
    } 
    \vspace{-6mm}
\label{fig:pre-dressing}
\end{figure}

Robotic-assisted dressing presents several challenges.
First of all, dressing is a task that entails the manipulation of a deformable object, which thus needs to deal with the infinite degrees of freedom of garments as well as self-occlusions~\cite{zhu2022challenges, longhini2024unfolding}.
Secondly, dressing assistance requires manipulation close to a human, imposing harder restrictions in the acceleration forces of the motions used compared to other deformable object manipulation tasks~\cite{ha2022flingbot}.
Although recent works have demonstrated great progress in assisted dressing~\cite{sun2024force, kotsovolis_2025_ral}, they commonly assume that the garment is already unfolded, ready for dressing.
In reality, however, garments such as medical gowns and aprons are typically stored in a folded configuration, requiring an additional unfolding step before they can be used.
For this reason, in this work we explicitly address what we denote as the \textit{pre-dressing} step: unfolding an initially folded garment and prepare it for assisted dressing, see Fig.~\ref{fig:pre-dressing}.

While recent research on deformable object manipulation tackles the problem of garment unfolding~\cite{ha2022flingbot,lin_2021_vcd,Huang_2022_mesh_based_dynamics, blancomulero_2023_qdp}, these works consider fabrics that start in a crumpled configuration.
To the best of our knowledge, this work is the first to address unfolding a garment starting from an initial folded configuration.
In this work, we investigate the effectiveness of different manipulation primitives for unfolding a medical gown. To manipulate the garment we leverage the imitation learning framework of \acp{dmp}~\cite{sidiropoulos2023novel}. Then, to evaluate the success of the task, we propose to use a visual classifier to determine the opening state of the garment. Supplementary materials, as well as details of the DMP and visual classifier, are available on our website\footnote{\url{https://sites.google.com/view/pre-dressing}\label{foot:website}}.

\begin{figure*}[ht]
\vspace{2mm}
    \centering
    \begin{subfigure}{0.13\textwidth}
        \centering
        \includegraphics[width=\linewidth]{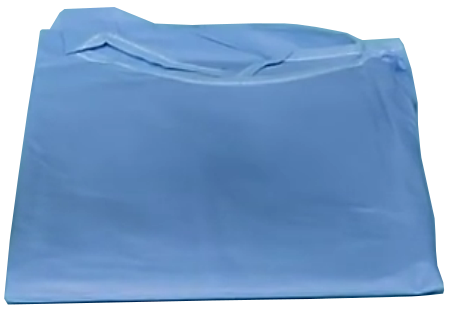}
        \caption{Folded.}
        \label{fig:subfig1}
    \end{subfigure}
    % \hfill
    \begin{subfigure}{0.23\textwidth}
        \centering
        \includegraphics[width=\linewidth]{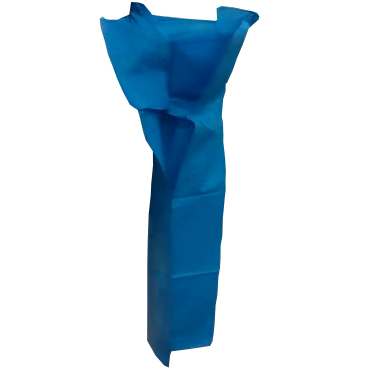}
        \caption{Closed with sleeves tangled.}
        \label{fig:subfig2}
    \end{subfigure}
    % \hfill
    \begin{subfigure}{0.155\textwidth}
        \centering
        \includegraphics[width=\linewidth]{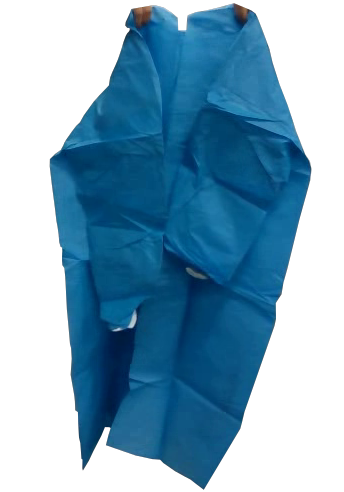}
        \caption{Closed untangled.}
        \label{fig:subfig3}
    \end{subfigure}
    % \vspace{1em} % Adjust spacing between rows
    \begin{subfigure}{0.165\textwidth}
        \centering
        \includegraphics[width=\linewidth]{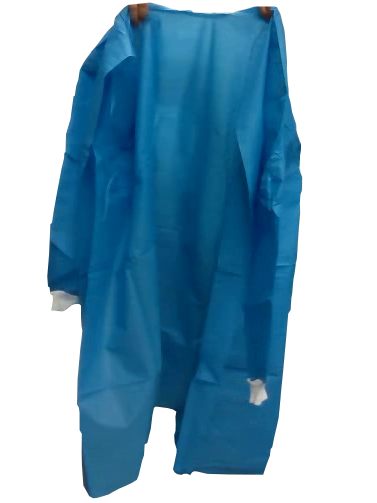}
        \caption{Partly Opened.}
        \label{fig:subfig4}
    \end{subfigure}
    % \hfill
    \begin{subfigure}{0.13\textwidth}
        \centering
        \includegraphics[width=\linewidth]{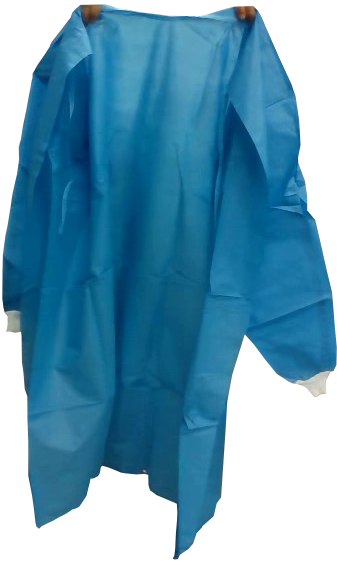}
        \caption{Opened 1.}
        \label{fig:subfig5}
    \end{subfigure}
    \begin{subfigure}{0.13\textwidth}
        \centering
        \includegraphics[width=\linewidth]{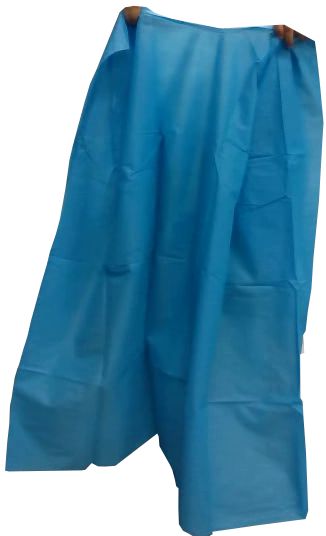}
        \caption{Opened 2.}
        \label{fig:subfig6}
    \end{subfigure}

    \caption{Example of a medical gown in: (a) folded configuration, and (b) closed configuration with the sleeves tangled after lifting the garment from a flat surface. The categories utilised for identifying the gown state and training a visual classifier are: (c) Closed, (d) Partly Opened, (e) Opened with sleeves forward, and (f) Opened with both sleeves hidden.}
    \label{fig:gown-states}
    \vspace{-2mm}
\end{figure*}

In summary, our contributions include:
\begin{itemize}
    \item Introducing the problem of pre-dressing, focusing on the unfolding of a garment before dressing assistance.
    \item Presenting a solution for pre-dressing based on \acp{dmp} and a visual classifier to identify the opening state of the garment.
    \item  Conducting an empirical evaluation of different manipulation primitives, comparing high-acceleration and low-acceleration manipulation actions to assess their effectiveness in unfolding garments. 
\end{itemize}

%% file: img_tex/main_v2.pdf_tex
%% Creator: Inkscape 1.3.2 (091e20e, 2023-11-25), www.inkscape.org
%% Accompanies image file 'main_v2.pdf' (pdf, eps, ps)
\begingroup%
  \makeatletter%
  \providecommand\color[2][]{%
    \errmessage{(Inkscape) Color is used for the text in Inkscape, but the package 'color.sty' is not loaded}%
    \renewcommand\color[2][]{}%
  }%
  \providecommand\transparent[1]{%
    \errmessage{(Inkscape) Transparency is used (non-zero) for the text in Inkscape, but the package 'transparent.sty' is not loaded}%
    \renewcommand\transparent[1]{}%
  }%
  \providecommand\rotatebox[2]{#2}%
  \newcommand*\fsize{\dimexpr\f@size pt\relax}%
  \newcommand*\lineheight[1]{\fontsize{\fsize}{#1\fsize}\selectfont}%
  \ifx\svgwidth\undefined%
    \setlength{\unitlength}{333.11930979bp}%
    \ifx\svgscale\undefined%
      \relax%
    \else%
      \setlength{\unitlength}{\unitlength * \real{\svgscale}}%
    \fi%
  \else%
    \setlength{\unitlength}{\svgwidth}%
  \fi%
  \global\let\svgwidth\undefined%
  \global\let\svgscale\undefined%
  \makeatother%
  \begin{picture}(1,0.86367605)%
    \lineheight{1}%
    \setlength\tabcolsep{0pt}%
    \put(0,0){\includegraphics[width=\unitlength,page=1]{main_v2.pdf}}%
    \put(0.1201191,0.82558269){\color[rgb]{0,0,0}\makebox(0,0)[lt]{\lineheight{1.25}\smash{\begin{tabular}[t]{l}1. Folded State\end{tabular}}}}%
    \put(0.60136799,0.82820131){\color[rgb]{0,0,0}\makebox(0,0)[lt]{\lineheight{1.25}\smash{\begin{tabular}[t]{l}2. Pre-Dressing Step\end{tabular}}}}%
    \put(0.13040289,0.50855175){\color[rgb]{0,0,0}\makebox(0,0)[lt]{\lineheight{1.25}\smash{\begin{tabular}[t]{l}3. Unfolded State\\Ready for Dressing\end{tabular}}}}%
    \put(0.64986034,0.40404303){\color[rgb]{0,0,0}\makebox(0,0)[t]{\lineheight{1.25}\smash{\begin{tabular}[t]{c}Constrained\\Learned\\Manipulation\end{tabular}}}}%
    \put(0,0){\includegraphics[width=\unitlength,page=2]{main_v2.pdf}}%
  \end{picture}%
\endgroup%

%% file: sections/related_works.tex
\subsection{Unfolding Deformable Objects}
The problem of unfolding a deformable object has been studied in the literature for objects such as cloths~\cite{blancomulero_2023_qdp}, garments~\cite{ha2022flingbot}, and bags~\cite{chen_2023_autobag}, to name a few.
Prior works on cloth and garment unfolding have primarily focused on the task of flattening~\cite{blancomulero_2023_qdp, ha2022flingbot,lin_2021_vcd,Huang_2022_mesh_based_dynamics,gu_2024_oriented_unfolding}.
In this task, given a garment in a crumpled state, the robot manipulator needs to find the necessary actions to flatten the object, maximising the area that it covers.
Similarly, recent works on bag unfolding address the problem of maximising the volume of an initial crumpled bag~\cite{chen_2023_autobag,hannus_2024_constrained_dmp_bags,chen2023SLIPbagging}.
These approaches share the common goal of moving the deformable object from an initially unstructured state to a more structured configuration.
In contrast, this work tackles the opposite problem, unfolding a garment from an initial structured folded configuration.

\subsection{Assisted Dressing}
Assisted dressing has gained significant attention over the past years~\cite{wang_2023_rss_onepolicy_dress, zhang_2022_assisted_dressing_science, sun2024force, zhu2024_doyouneedahand_bimanual_dressing_mbmf, kotsovolis_2025_ral}.
Here, we primarily focus on the recent advances related to dressing the upper body of a human, one important aspect in occupational therapy. 
Previous works have developed learning-based approaches for inserting the garment sleeves onto a person's arm~\cite{sun2024force, kotsovolis_2025_ral}, assuming the garment is already grasped and ready for dressing.
Alternatively, other works have also incorporated grasping the garment into their assisted dressing pipeline~\cite{zhang_2022_assisted_dressing_science, zhang_2023_visuotactile_garment_unfold}.
As an example, Zhang et al.~\cite{zhang_2023_visuotactile_garment_unfold} proposed an edge tracing method for preparing the garment before dressing.
However, the aforementioned works assume that  the garment is already unfolded, facilitating the dressing step.
In this work, we tackle this crucial step required prior to the dressing assistance step, evaluating the effectiveness of different manipulation primitives to unfold the garment.

%% file: sections/method.tex
\subsection{Task Definition}
\label{sec:method-task-def}
The objective of the pre-dressing task is to prepare a folded garment for dressing assistance.
Hence, given a garment in a folded configuration placed on a flat surface, the robot needs to perform the necessary actions to unfold the garment (see Fig.~\ref{fig:pre-dressing}).
More specifically, in this paper we focus on the task of unfolding medical gowns.
Unlike other garments such as t-shirts, a folded gown may exhibit sleeve intersections, adding complexity to its unfolding process.
As shown in Fig.~\ref{fig:subfig2}, when the gown is grasped from a flat surface and the contact is removed, the overlapped fabric layers hinder the complete opening of the garment.
In line with prior works on robotic-assisted dressing~\cite{zhang_2021_1d_dynamic, sun2024force, zhang_2023_visuotactile_garment_unfold}, we consider a bi-manual robot set-up scenario.
In addition, we assume that the detection of the grasping location is solved, and perform a top grasp of a single layer of the garment in the collar.

\subsection{Metrics for Evaluating Unfolding Success}
\label{sec:method-metrics}
% To measure the success of ... 
Following the definition of bending levels proposed in the taxonomy in~\cite{blancomulero2024tdom}, the pre-dressing task requires to move from an structured (folded) configuration to an unstructured (opened) bending configuration, increasing the unstructuredness of the garment.
The unstructured configuration is defined by the number of accessible keypoints.
Here, once the garment has been lifted and the sleeves are untangled, losing visibility of the sleeves indicates that the opening state has improved.
Hence, we utilise as keypoints the position of the sleeves.

To identify the state of the garment, rather than manually engineering a reward based on the position of the sleeves, we train a visual classifier to identify the opening state.
We define three categories, illustrated in Fig.~\ref{fig:gown-states}, as follows:
\begin{itemize}
    \item \textbf{Closed:} the sleeves of the gown are either tangled and not visible or fully visible at the back of the gown, see Fig.~\ref{fig:subfig2} and Fig.~\ref{fig:subfig3}. Furthermore, the left and right back sides of the gown are tangled.
    \item \textbf{Partly Opened:} at least one of the sleeves faces forward.
    In addition, the back sides of the gown are untangled and slightly opened, see Fig.~\ref{fig:subfig4}.
    \item \textbf{Opened:} one or both sleeves are either partly visible or not visible. Here, both back sides of the gown are untangled and widely opened, see Fig.~\ref{fig:subfig5} and Fig.~\ref{fig:subfig6}. 
\end{itemize}
Note that the main difference between the categories of partly opened and opened lies in the position of the sleeves and a wider opening of the back side of the gown.

The approach of recognising the state using a visual classifier circumvents the challenge of self-occlusion, which arises when using motion capture systems.
Unlike other deformable objects such as bags~\cite{hannus_2024_constrained_dmp_bags}, where positioning markers remain visible,  markers placed on the sleeves of a gown may become obscured during unfolding.
Another alternative would be to utilise the Chamfer Distance, which has been previously used for identifying the success in cloth unfolding tasks~\cite{lin_2021_vcd, Huang_2022_mesh_based_dynamics}.
However, this would require access to a canonical shape. 
In this task, there are thousands of valid configurations for pre-dressing, which would require a vast amount of canonical shapes to contrast against, undermining the potential use of such metric.

\subsection{Manipulation Primitives for Pre-Dressing}
\label{sec:method-manip-primitives}
To comprehensively study the complexity of the pre-dressing step, we devise three bi-manual manipulation primitives that target different unfolding dynamics of the gown. The manipulation primitives are as follows:
\begin{itemize}
    \item \textbf{Fling}: similar to the primitives used in~\cite{ha2022flingbot,hannus_2024_constrained_dmp_bags}, this dynamic motion entails rapidly accelerating the robot, leveraging inertia to push air into the gown and improve its opening state. The main rotation takes place around the Y-axis of the robot frame.
    \item \textbf{Shake}: rapid motion where the robot moves repeatedly back and forth while rotating the end-effector to loosen the garment folds. As in the previous motion, the main rotation is around the Y-axis of the robot frame.
    \item \textbf{Twist}: motion with lower accelerations where the robot moves twice forward while rotating the end-effector in opposite directions to create tension in the garment. In this motion the main rotation is around the Z-axis of the robot frame.
\end{itemize}

One solution for learning these primitives is utilising Reinforcement Learning and training a policy in simulation~\cite{blancomulero_2023_qdp, ha2022flingbot}.
However, due to the sim-to-real gap in garment simulation~\cite{blancomulero_2024_benchmarking_cloth_manip}, as well as the challenge of simulating the aerodynamics when dynamically unfolding a deformable object~\cite{hannus_2024_constrained_dmp_bags}, we decide to apply the imitation learning framework of \acp{dmp}~\cite{sidiropoulos2023novel} to circumvent these challenges. 

The three motions are demonstrated by a human and captured using a motion capture system.
Then, the demonstrations are pre-processed to reduce noise from the capture system and to filter out movements along non-essential axes.
Additionally, the maximum distance between the end-effectors is constrained to prevent excessive tension that could tear apart the garment.
Taking into account that a learned dynamic motion can exceed the hardware limitations of the manipulator, we follow the same approach as in~\cite{hannus_2024_constrained_dmp_bags}
and use a \ac{dmp} formulation that constrains the position, velocity and acceleration to those of the robot system~\cite{sidiropoulos2023novel}.

In addition to the three learned manipulation primitives, we introduce a quasi-static primitive inspired by~\cite{hannus_2024_constrained_dmp_bags}, designed to further refine the opening state of the garment.
This motion slowly moves forward the garment, maintaining the distance between the end-effectors, and finally positioning the gown for handover to a human for the dressing step.

%% file: img_tex/exp_prims_v2.pdf_tex
%% Creator: Inkscape 1.3.2 (091e20e, 2023-11-25), www.inkscape.org
%% Accompanies image file 'exp_prims_v2.pdf' (pdf, eps, ps)
\begingroup%
  \makeatletter%
  \providecommand\color[2][]{%
    \errmessage{(Inkscape) Color is used for the text in Inkscape, but the package 'color.sty' is not loaded}%
    \renewcommand\color[2][]{}%
  }%
  \providecommand\transparent[1]{%
    \errmessage{(Inkscape) Transparency is used (non-zero) for the text in Inkscape, but the package 'transparent.sty' is not loaded}%
    \renewcommand\transparent[1]{}%
  }%
  \providecommand\rotatebox[2]{#2}%
  \newcommand*\fsize{\dimexpr\f@size pt\relax}%
  \newcommand*\lineheight[1]{\fontsize{\fsize}{#1\fsize}\selectfont}%
  \ifx\svgwidth\undefined%
    \setlength{\unitlength}{630.54287167bp}%
    \ifx\svgscale\undefined%
      \relax%
    \else%
      \setlength{\unitlength}{\unitlength * \real{\svgscale}}%
    \fi%
  \else%
    \setlength{\unitlength}{\svgwidth}%
  \fi%
  \global\let\svgwidth\undefined%
  \global\let\svgscale\undefined%
  \makeatother%
  \begin{picture}(1,0.64266792)%
    \lineheight{1}%
    \setlength\tabcolsep{0pt}%
    \put(0,0){\includegraphics[width=\unitlength,page=1]{exp_prims_v2.pdf}}%
    \put(0.76493089,0.62536169){\color[rgb]{0,0,0}\makebox(0,0)[t]{\lineheight{1.25}\smash{\begin{tabular}[t]{c}\textbf{Packed Gowns}\end{tabular}}}}%
    \put(0.26003332,0.62536169){\color[rgb]{0,0,0}\makebox(0,0)[t]{\lineheight{1.25}\smash{\begin{tabular}[t]{c}\textbf{Previously Opened Gowns}\end{tabular}}}}%
    \put(0.60011365,0.60202415){\color[rgb]{0,0,0}\makebox(0,0)[t]{\lineheight{1.25}\smash{\begin{tabular}[t]{c}Fling\end{tabular}}}}%
    \put(0.76340227,0.60202415){\color[rgb]{0,0,0}\makebox(0,0)[t]{\lineheight{1.25}\smash{\begin{tabular}[t]{c}Shake\end{tabular}}}}%
    \put(0.92718969,0.60202415){\color[rgb]{0,0,0}\makebox(0,0)[t]{\lineheight{1.25}\smash{\begin{tabular}[t]{c}Twist\end{tabular}}}}%
    \put(0.09521605,0.60202422){\color[rgb]{0,0,0}\makebox(0,0)[t]{\lineheight{1.25}\smash{\begin{tabular}[t]{c}Fling\end{tabular}}}}%
    \put(0.04382805,0.32064693){\color[rgb]{0,0,0}\makebox(0,0)[lt]{\lineheight{1.25}\smash{\begin{tabular}[t]{l}Opened (82\%)\end{tabular}}}}%
    \put(0.04382805,0.03190289){\color[rgb]{0,0,0}\makebox(0,0)[lt]{\lineheight{1.25}\smash{\begin{tabular}[t]{l}Opened (73\%)\\Iteration \# 1\end{tabular}}}}%
    \put(0.20717631,0.03190289){\color[rgb]{0,0,0}\makebox(0,0)[lt]{\lineheight{1.25}\smash{\begin{tabular}[t]{l}Opened (88\%)\\Iteration \# 1\end{tabular}}}}%
    \put(0.37052462,0.03190289){\color[rgb]{0,0,0}\makebox(0,0)[lt]{\lineheight{1.25}\smash{\begin{tabular}[t]{l}Opened (90\%)\\Iteration \# 1\end{tabular}}}}%
    \put(0.55213255,0.03190289){\color[rgb]{0,0,0}\makebox(0,0)[lt]{\lineheight{1.25}\smash{\begin{tabular}[t]{l}Partly (93\%)\\Iteration \# 4\end{tabular}}}}%
    \put(0.71571737,0.03190289){\color[rgb]{0,0,0}\makebox(0,0)[lt]{\lineheight{1.25}\smash{\begin{tabular}[t]{l}Partly (91\%)\\Iteration \# 1\end{tabular}}}}%
    \put(0.87927821,0.03190289){\color[rgb]{0,0,0}\makebox(0,0)[lt]{\lineheight{1.25}\smash{\begin{tabular}[t]{l}Partly (95\%)\\Iteration \# 2\end{tabular}}}}%
    \put(0.2585047,0.60202422){\color[rgb]{0,0,0}\makebox(0,0)[t]{\lineheight{1.25}\smash{\begin{tabular}[t]{c}Shake\end{tabular}}}}%
    \put(0.42229208,0.60202422){\color[rgb]{0,0,0}\makebox(0,0)[t]{\lineheight{1.25}\smash{\begin{tabular}[t]{c}Twist\end{tabular}}}}%
    \put(0.00871529,0.00188875){\color[rgb]{0,0,0}\makebox(0,0)[t]{\lineheight{1.25}\smash{\begin{tabular}[t]{c}(a)\end{tabular}}}}%
    \put(0.5121768,0.00188175){\color[rgb]{0,0,0}\makebox(0,0)[t]{\lineheight{1.25}\smash{\begin{tabular}[t]{c}(b)\end{tabular}}}}%
    \put(0,0){\includegraphics[width=\unitlength,page=2]{exp_prims_v2.pdf}}%
    \put(0.21456392,0.32064693){\color[rgb]{0,0,0}\makebox(0,0)[lt]{\lineheight{1.25}\smash{\begin{tabular}[t]{l}Partly (94\%)\end{tabular}}}}%
    \put(0.37453553,0.3201033){\color[rgb]{0,0,0}\makebox(0,0)[lt]{\lineheight{1.25}\smash{\begin{tabular}[t]{l}Closed (82\%)\end{tabular}}}}%
    \put(0.55270753,0.3201033){\color[rgb]{0,0,0}\makebox(0,0)[lt]{\lineheight{1.25}\smash{\begin{tabular}[t]{l}Closed (91\%)\end{tabular}}}}%
    \put(0.71941117,0.32064693){\color[rgb]{0,0,0}\makebox(0,0)[lt]{\lineheight{1.25}\smash{\begin{tabular}[t]{l}Partly (88\%)\end{tabular}}}}%
    \put(0.87953956,0.3201033){\color[rgb]{0,0,0}\makebox(0,0)[lt]{\lineheight{1.25}\smash{\begin{tabular}[t]{l}Closed (91\%)\end{tabular}}}}%
    \put(0,0){\includegraphics[width=\unitlength,page=3]{exp_prims_v2.pdf}}%
  \end{picture}%
\endgroup%

%% file: sections/experiment_setup.tex
The goal of our experiments is to assess the requirements for succeeding in the pre-dressing step.
To that aim, our experiments investigate: 1) which manipulation primitives are more suitable for pre-dressing, 2) the performance of combining motion primitives, and 3) the performance in previously opened and recently unpacked gowns.

\subsection{Experimental Set-up}
Our set-up consists of a dual-arm system of two 6-DoF UR5e robot arms, which are mounted in a platform resembling the pose of a human.
The robot arms are equipped with a gripper designed for the EU project SoftEnable \cite{foix2024Gripper}, which enables automatically grasping a single layer of the garment.
Our set-up also includes a RealSense D435i camera for capturing the gown configuration and classifying its state.
Additionally, a table is placed beneath the robots for grasping the flat gowns. The table is set at a height ensuring that the garment remains suspended in the air after being lifted.

% All the garments are positioned in a folded configuration.
Our experiments evaluate garments in two configurations: 1) medical gowns that have been previously opened, and 2) recently unpackaged medical gowns. 
For the previously opened gowns, we use the same gown, folding it in the same sequence in which the sleeves are tangled, ending after multiple folds in a square folded configuration.
The unpacked gowns start from a randomly folded configuration as they come out of a sealed bag.
Thus, some of these garments may exhibit more complex folds that require more effort to unfold.
In the experiments, the initial configuration of the gown once lifted from the flat surface is random. 
Each experiment consists of up to 5 iterations to assess improvement in gown opening.
For previously opened gowns we performed three trials per primitive. 
For packed gowns, due to time and resource constraints, we performed one trial per primitive or combination, except for primitive and quasi-static combination, in which we performed two trials.

For the constrained learned \acp{dmp} we set the same constraints as in ~\cite{hannus_2024_constrained_dmp_bags}, that is, 98\% of the joint position, velocity and acceleration limits.
Finally, for the visual classifier we utilise YOLOv11~\cite{Jocher_Ultralytics_YOLO_2023} and train it using a dataset created from samples of the three categories presented in Section~\ref{sec:method-metrics}, gathering data from manipulation performed by a human.
Additional details can be found on our website\footref{foot:website}.

%% file: sections/results.tex
\subsection{Manipulation Primitives Comparison} 
\label{sec: fling_only_experiment}
We start by evaluating the three learned primitives without the quasi-static motion.
The results are shown in Table~\ref{tab:prev-opened-results} and Fig.~\ref{fig:exp-primitives}.
The percentage in Table~\ref{tab:prev-opened-results} indicates the percentage of experiments in which the gown was classified as opened or as partly opened.
First, we analyse the results of previously opened gowns.
We can notice that all primitives achieve a large success rate, where the twist primitive performs best. 
Additionally, all the primitives manage to move the sleeves to the forward position of the gown.
However, looking at Fig.~\ref{fig:exp-primitives}, the only primitive that results in the sleeves not visible by the camera is the twist primitive.
This is a result of the final position of the robot arms in the twist primitive, which rotates the grippers facilitating the opened state.
Moreover, all primitives take on average only one iteration to achieve a fully opened state.
In general, this indicates that both quasi-static and dynamic primitives are great candidates for unfolding a garment which has been previously opened.

Now, looking at the results for packed gowns we observe that the performance drastically drops. 
None of the primitives is able to reach the opened state.
Furthermore, while all primitives are able to achieve a partly opened state, the number of iterations increases.
It is important to note that although the shake primitive is able to partly open the gown within one iteration, the opening is significantly worse than the one achieved by the twist primitive, as shown in Fig.~\ref{fig:exp-primitives}.
By contrast, the width of the gown after applying either the fling or the twist is significantly larger, while requiring more iterations.
These results highlight the need for a motion that refines the state of the gown further.

% Separate file for the table 1 to move it more easily
\input{sections/table_1_prims}
\begin{figure*}
\centering
\def\svgwidth{\linewidth}
{
\vspace{2mm}
    \fontsize{7}{7}%\selectfont\sf
    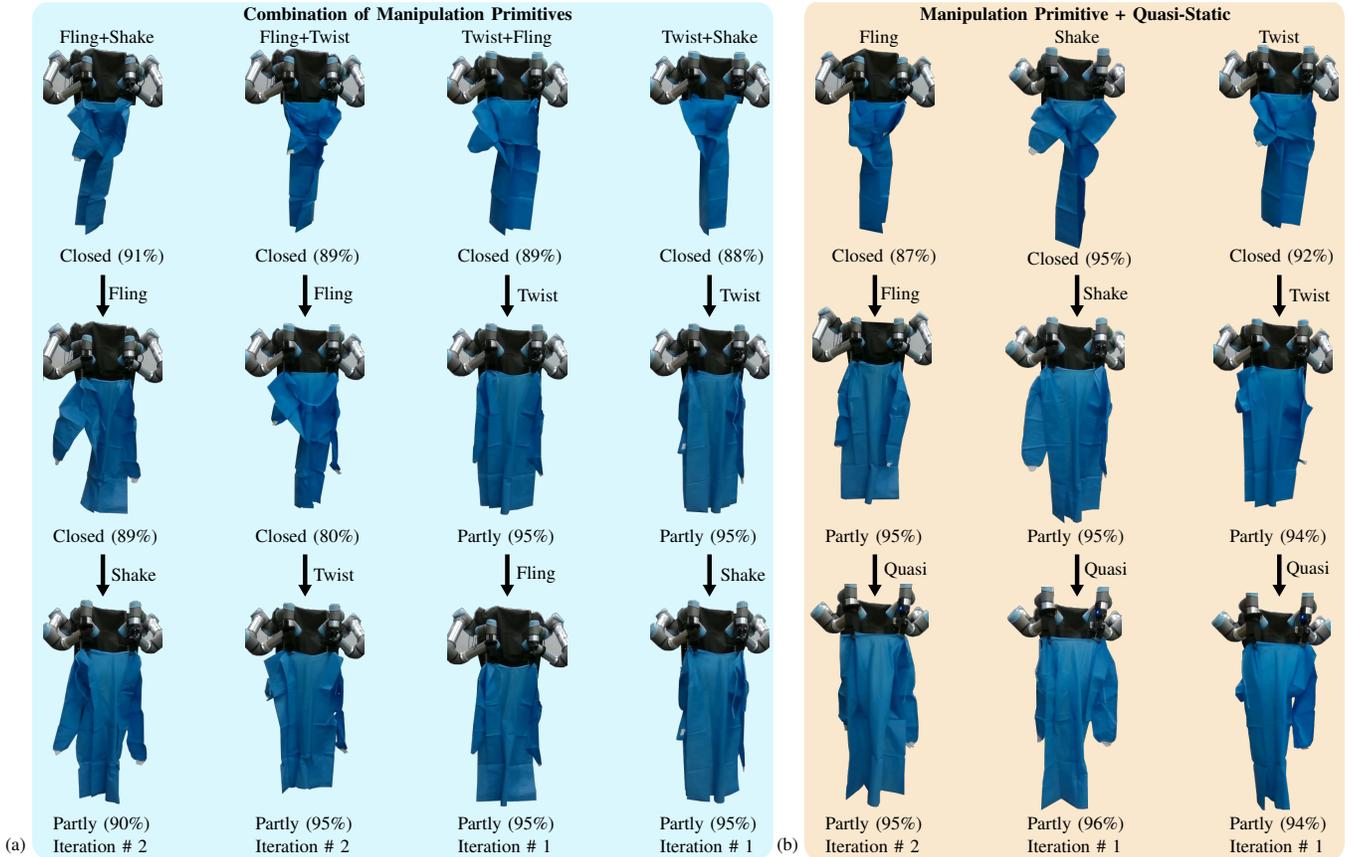}
    \caption{Qualitative results for unfolding unpacked gowns combining the three manipulation primitives (a) with each other and (b) with a quasi-static motion. The first row shows the gown shape before execution, while the second and third rows show the resulting gown state after the corresponding manipulation indicated next to an arrow.
    We indicate the category predicted by the visual classifier and its confidence, as well as the iteration at which a partly opened state was reached.
    % where the original shape and the deformation are indicated in black and red colours, respectively. 
    } 
\label{fig:exp-primitives-comb}
\vspace{-1mm}
\end{figure*}

\subsection{Combining Dynamic and Quasi-Static Motions}
\label{sec: fling_only_experiment}
Next, we investigate the effectiveness of combining different motions for the unpacked gown case.
Here, we evaluate two approaches: 1) combining two of the learned manipulation primitives, and 2) performing the quasi-static motion  described in Section~\ref{sec:method-manip-primitives} after a learned manipulation primitive.
The results are shown in Table~\ref{tab:results-prims-combined} and Fig.~\ref{fig:exp-primitives-comb}.

Firstly, we analyse the results from the combination of the learned primitives.
We select as starting manipulations the twist and fling primitives, where the first is selected due to the superior performance in the prior results, and the second to evaluate the performance of the fling dynamic motion followed by primitives with lower accelerations.
In this experiment, an iteration involves the execution of two primitives. The combination of the fling with either the shake or twist motion requires an average of two iterations.
This yields the same number of motions as the fling standalone motion, four in total.
Secondly, we analyse the results of combining the learned primitives with the quasi-static motion.
Compared to the results in Table~\ref{tab:prev-opened-results}, the quasi-static motion helps to reduce the number of iterations required for partly opening the gown.
These results showcase that combining different motion primitives helps to unfold the garment further than single primitives for unpacked gowns. 

More generally, looking at Table~\ref{tab:results-prims-combined} we can notice that none of the combined approaches is able to achieve a fully opened state.
This results from the fact that the electrostatic forces between the layers of the unpacked garment are really high, and even after using dynamic manipulation primitives, the manipulation does not separate these layers.
Nevertheless, looking at Fig.~\ref{fig:exp-primitives-comb}, the partly opened state is such that the back of the gown is opened enough for a human to insert the arms.
Furthermore, the sleeves are positioned at the front, where all manipulations achieve 100\% success, which enables further manipulation for the subsequent dressing step. 

% Separate file for the table 2 to move it more easily
\input{sections/table_2_comb}

%% file: sections/table_1_prims.tex
\begin{table}[b!]
% \vspace{2mm}
\vspace{-4mm}
    \caption{Quantitative results for previously opened and unpacked gowns
    using the three learned primitives. The results indicate the percentage of gowns classified as opened or partly opened according to the trained classifier, and if the arms are positioned Forward (Fwd.), which indicates that the gown is fully opened. The results also indicate the mean number of iterations (It.) that it takes to reach either the open or partly opened state of the gown.} 
    \resizebox{\linewidth}{!}{%
        {\renewcommand\arraystretch{1.5} 
            \centering
            \begin{tabular}{m{0.07\linewidth}m{0.1\linewidth}|m{0.1\linewidth}|m{0.1\linewidth}|m{0.08\linewidth}|m{0.1\linewidth}}
                % \Cline{1pt}{3-6}
                \multicolumn{2}{c}{ }
                & \multicolumn{1}{|c|}{\parbox[t]{9mm}{\centering Opened}} 
                & \multicolumn{1}{c|}{\parbox[t]{9mm}{\centering Partly Opened}}
                & \multicolumn{1}{c|}{\parbox[t]{9mm}{\centering Arms Fwd.}}
                & \multicolumn{1}{c}{\parbox[t]{5mm}{\centering \# It.}}\\
                \noalign{\hrule height 1pt}
                % Prev Opened
                \centering\parbox[t]{5mm}{\multirow{3}{*}{\rotatebox[origin=c]{90}{{\parbox{1cm}{\textbf{Prev. Opened}}}}}} 
                & \multicolumn{1}{|c|}{Fling} 
                & \multicolumn{1}{c|}{33.33\%} & \multicolumn{1}{c|}{66.67\%} 
                & \multicolumn{1}{c|}{100\%} 
                & \multicolumn{1}{c}{1} \\ \cline{2-6}
                % Row 2
                & \multicolumn{1}{|c|}{Shake} 
                & \multicolumn{1}{c|}{66.67\%} & \multicolumn{1}{c|}{33.33\%} 
                & \multicolumn{1}{c|}{100\%} 
                & \multicolumn{1}{c}{1}  \\ \cline{2-6}
                % Row 3
                & \multicolumn{1}{|c|}{Twist} 
                & \multicolumn{1}{c|}{100\%} & \multicolumn{1}{c|}{0\%}
                & \multicolumn{1}{c|}{100\%} 
                & \multicolumn{1}{c}{1}  \\ 
                \noalign{\hrule height 1pt}
                % Packed gowns
                \centering\parbox[t]{1mm}{\multirow{3}{*}{\rotatebox[origin=c]{90}{{\textbf{Unpacked}}}}} 
                & \multicolumn{1}{|c|}{Fling} 
                & \multicolumn{1}{c|}{0\%} & \multicolumn{1}{c|}{100\%} 
                & \multicolumn{1}{c|}{0\%}  
                & \multicolumn{1}{c}{4} \\ \cline{2-6}
                % Row 2
                & \multicolumn{1}{|c|}{Shake} 
                & \multicolumn{1}{c|}{0\%} & \multicolumn{1}{c|}{100\%}
                & \multicolumn{1}{c|}{0\%} 
                & \multicolumn{1}{c}{1}  \\ \cline{2-6}
                % Row 3
                & \multicolumn{1}{|c|}{Twist}
                & \multicolumn{1}{c|}{0\%} & \multicolumn{1}{c|}{100\%}
                & \multicolumn{1}{c|}{0\%}
                & \multicolumn{1}{c}{2}  \\
                \noalign{\hrule height 1pt}
            \end{tabular}
        }}
    \label{tab:prev-opened-results}
\end{table}

%% file: img_tex/exp_prims_comb_v2.pdf_tex
%% Creator: Inkscape 1.3.2 (091e20e, 2023-11-25), www.inkscape.org
%% Accompanies image file 'exp_prims_comb_v2.pdf' (pdf, eps, ps)
\begingroup%
  \makeatletter%
  \providecommand\color[2][]{%
    \errmessage{(Inkscape) Color is used for the text in Inkscape, but the package 'color.sty' is not loaded}%
    \renewcommand\color[2][]{}%
  }%
  \providecommand\transparent[1]{%
    \errmessage{(Inkscape) Transparency is used (non-zero) for the text in Inkscape, but the package 'transparent.sty' is not loaded}%
    \renewcommand\transparent[1]{}%
  }%
  \providecommand\rotatebox[2]{#2}%
  \newcommand*\fsize{\dimexpr\f@size pt\relax}%
  \newcommand*\lineheight[1]{\fontsize{\fsize}{#1\fsize}\selectfont}%
  \ifx\svgwidth\undefined%
    \setlength{\unitlength}{681.23975859bp}%
    \ifx\svgscale\undefined%
      \relax%
    \else%
      \setlength{\unitlength}{\unitlength * \real{\svgscale}}%
    \fi%
  \else%
    \setlength{\unitlength}{\svgwidth}%
  \fi%
  \global\let\svgwidth\undefined%
  \global\let\svgscale\undefined%
  \makeatother%
  \begin{picture}(1,0.64800072)%
    \lineheight{1}%
    \setlength\tabcolsep{0pt}%
    \put(0,0){\includegraphics[width=\unitlength,page=1]{exp_prims_comb_v2.pdf}}%
    \put(0.79853031,0.62798098){\color[rgb]{0,0,0}\makebox(0,0)[t]{\lineheight{1.25}\smash{\begin{tabular}[t]{c}\textbf{Manipulation Primitive + Quasi-Static}\end{tabular}}}}%
    \put(0,0){\includegraphics[width=\unitlength,page=2]{exp_prims_comb_v2.pdf}}%
    \put(0.29955057,0.62798098){\color[rgb]{0,0,0}\makebox(0,0)[t]{\lineheight{1.25}\smash{\begin{tabular}[t]{c}\textbf{Combination of Manipulation Primitives}\end{tabular}}}}%
    \put(0.2225948,0.61078399){\color[rgb]{0,0,0}\makebox(0,0)[t]{\lineheight{1.25}\smash{\begin{tabular}[t]{c}Fling+Twist\end{tabular}}}}%
    \put(0.07462893,0.61078399){\color[rgb]{0,0,0}\makebox(0,0)[t]{\lineheight{1.25}\smash{\begin{tabular}[t]{c}Fling+Shake\end{tabular}}}}%
    \put(0.09046274,0.41823439){\color[rgb]{0,0,0}\makebox(0,0)[t]{\lineheight{1.25}\smash{\begin{tabular}[t]{c}Fling\end{tabular}}}}%
    \put(0.24400656,0.41823439){\color[rgb]{0,0,0}\makebox(0,0)[t]{\lineheight{1.25}\smash{\begin{tabular}[t]{c}Fling\end{tabular}}}}%
    \put(0.39534845,0.20913321){\color[rgb]{0,0,0}\makebox(0,0)[t]{\lineheight{1.25}\smash{\begin{tabular}[t]{c}Fling\end{tabular}}}}%
    \put(0.37373174,0.61078399){\color[rgb]{0,0,0}\makebox(0,0)[t]{\lineheight{1.25}\smash{\begin{tabular}[t]{c}Twist+Fling\end{tabular}}}}%
    \put(0.52533031,0.61078399){\color[rgb]{0,0,0}\makebox(0,0)[t]{\lineheight{1.25}\smash{\begin{tabular}[t]{c}Twist+Shake\end{tabular}}}}%
    \put(0.0062741,0.00620333){\color[rgb]{0,0,0}\makebox(0,0)[t]{\lineheight{1.25}\smash{\begin{tabular}[t]{c}(a)\end{tabular}}}}%
    \put(0,0){\includegraphics[width=\unitlength,page=3]{exp_prims_comb_v2.pdf}}%
    \put(0.0946035,0.20813728){\color[rgb]{0,0,0}\makebox(0,0)[t]{\lineheight{1.25}\smash{\begin{tabular}[t]{c}Shake\end{tabular}}}}%
    \put(0.18561752,0.23695254){\color[rgb]{0,0,0}\makebox(0,0)[lt]{\lineheight{1.25}\smash{\begin{tabular}[t]{l}Closed (80\%)\end{tabular}}}}%
    \put(0.33680966,0.23695254){\color[rgb]{0,0,0}\makebox(0,0)[lt]{\lineheight{1.25}\smash{\begin{tabular}[t]{l}Partly (95\%)\\\end{tabular}}}}%
    \put(0.33680966,0.4465445){\color[rgb]{0,0,0}\makebox(0,0)[lt]{\lineheight{1.25}\smash{\begin{tabular}[t]{l}Closed (89\%)\end{tabular}}}}%
    \put(0.48800182,0.4465445){\color[rgb]{0,0,0}\makebox(0,0)[lt]{\lineheight{1.25}\smash{\begin{tabular}[t]{l}Closed (88\%)\end{tabular}}}}%
    \put(0.48800182,0.23695254){\color[rgb]{0,0,0}\makebox(0,0)[lt]{\lineheight{1.25}\smash{\begin{tabular}[t]{l}Partly (95\%)\\\end{tabular}}}}%
    \put(0.55031286,0.20813728){\color[rgb]{0,0,0}\makebox(0,0)[t]{\lineheight{1.25}\smash{\begin{tabular}[t]{c}Shake\end{tabular}}}}%
    \put(0.24422516,0.20813728){\color[rgb]{0,0,0}\makebox(0,0)[t]{\lineheight{1.25}\smash{\begin{tabular}[t]{c}Twist\end{tabular}}}}%
    \put(0.39666799,0.41723846){\color[rgb]{0,0,0}\makebox(0,0)[t]{\lineheight{1.25}\smash{\begin{tabular}[t]{c}Twist\end{tabular}}}}%
    \put(0.54800994,0.41723846){\color[rgb]{0,0,0}\makebox(0,0)[t]{\lineheight{1.25}\smash{\begin{tabular}[t]{c}Twist\end{tabular}}}}%
    \put(0,0){\includegraphics[width=\unitlength,page=4]{exp_prims_comb_v2.pdf}}%
    \put(0.03442542,0.02207248){\color[rgb]{0,0,0}\makebox(0,0)[lt]{\lineheight{1.25}\smash{\begin{tabular}[t]{l}Partly (90\%)\\Iteration \# 2\end{tabular}}}}%
    \put(0.18561754,0.02207248){\color[rgb]{0,0,0}\makebox(0,0)[lt]{\lineheight{1.25}\smash{\begin{tabular}[t]{l}Partly (95\%)\\Iteration \# 2\end{tabular}}}}%
    \put(0.33680966,0.02207248){\color[rgb]{0,0,0}\makebox(0,0)[lt]{\lineheight{1.25}\smash{\begin{tabular}[t]{l}Partly (95\%)\\Iteration \# 1\end{tabular}}}}%
    \put(0.48800182,0.02207248){\color[rgb]{0,0,0}\makebox(0,0)[lt]{\lineheight{1.25}\smash{\begin{tabular}[t]{l}Partly (95\%)\\Iteration \# 1\end{tabular}}}}%
    \put(0.03442542,0.23695254){\color[rgb]{0,0,0}\makebox(0,0)[lt]{\lineheight{1.25}\smash{\begin{tabular}[t]{l}Closed (89\%)\end{tabular}}}}%
    \put(0.03954792,0.4465445){\color[rgb]{0,0,0}\makebox(0,0)[lt]{\lineheight{1.25}\smash{\begin{tabular}[t]{l}Closed (91\%)\end{tabular}}}}%
    \put(0.18561752,0.4465445){\color[rgb]{0,0,0}\makebox(0,0)[lt]{\lineheight{1.25}\smash{\begin{tabular}[t]{l}Closed (89\%)\end{tabular}}}}%
    \put(0,0){\includegraphics[width=\unitlength,page=5]{exp_prims_comb_v2.pdf}}%
    \put(0.79983018,0.61078399){\color[rgb]{0,0,0}\makebox(0,0)[t]{\lineheight{1.25}\smash{\begin{tabular}[t]{c}Shake\end{tabular}}}}%
    \put(0.65186428,0.61078399){\color[rgb]{0,0,0}\makebox(0,0)[t]{\lineheight{1.25}\smash{\begin{tabular}[t]{c}Fling\end{tabular}}}}%
    \put(0.6676981,0.41823439){\color[rgb]{0,0,0}\makebox(0,0)[t]{\lineheight{1.25}\smash{\begin{tabular}[t]{c}Fling\end{tabular}}}}%
    \put(0.82124193,0.41823439){\color[rgb]{0,0,0}\makebox(0,0)[t]{\lineheight{1.25}\smash{\begin{tabular}[t]{c}Shake\end{tabular}}}}%
    \put(0.97258382,0.21353695){\color[rgb]{0,0,0}\makebox(0,0)[t]{\lineheight{1.25}\smash{\begin{tabular}[t]{c}Quasi\end{tabular}}}}%
    \put(0.95096709,0.61078399){\color[rgb]{0,0,0}\makebox(0,0)[t]{\lineheight{1.25}\smash{\begin{tabular}[t]{c}Twist\end{tabular}}}}%
    \put(0.58350947,0.00620333){\color[rgb]{0,0,0}\makebox(0,0)[t]{\lineheight{1.25}\smash{\begin{tabular}[t]{c}(b)\end{tabular}}}}%
    \put(0,0){\includegraphics[width=\unitlength,page=6]{exp_prims_comb_v2.pdf}}%
    \put(0.67183886,0.21254102){\color[rgb]{0,0,0}\makebox(0,0)[t]{\lineheight{1.25}\smash{\begin{tabular}[t]{c}Quasi\end{tabular}}}}%
    \put(0.76285286,0.23695254){\color[rgb]{0,0,0}\makebox(0,0)[lt]{\lineheight{1.25}\smash{\begin{tabular}[t]{l}Partly (95\%)\end{tabular}}}}%
    \put(0.91404504,0.23695254){\color[rgb]{0,0,0}\makebox(0,0)[lt]{\lineheight{1.25}\smash{\begin{tabular}[t]{l}Partly (94\%)\\\end{tabular}}}}%
    \put(0.91404504,0.4465445){\color[rgb]{0,0,0}\makebox(0,0)[lt]{\lineheight{1.25}\smash{\begin{tabular}[t]{l}Closed (92\%)\end{tabular}}}}%
    \put(0.82146053,0.21254102){\color[rgb]{0,0,0}\makebox(0,0)[t]{\lineheight{1.25}\smash{\begin{tabular}[t]{c}Quasi\end{tabular}}}}%
    \put(0.97390338,0.41723846){\color[rgb]{0,0,0}\makebox(0,0)[t]{\lineheight{1.25}\smash{\begin{tabular}[t]{c}Twist\end{tabular}}}}%
    \put(0.61166078,0.02207248){\color[rgb]{0,0,0}\makebox(0,0)[lt]{\lineheight{1.25}\smash{\begin{tabular}[t]{l}Partly (95\%)\\Iteration \# 2\end{tabular}}}}%
    \put(0.76285286,0.02207248){\color[rgb]{0,0,0}\makebox(0,0)[lt]{\lineheight{1.25}\smash{\begin{tabular}[t]{l}Partly (96\%)\\Iteration \# 1\end{tabular}}}}%
    \put(0.91404504,0.02207248){\color[rgb]{0,0,0}\makebox(0,0)[lt]{\lineheight{1.25}\smash{\begin{tabular}[t]{l}Partly (94\%)\\Iteration \# 1\end{tabular}}}}%
    \put(0.61166078,0.23695254){\color[rgb]{0,0,0}\makebox(0,0)[lt]{\lineheight{1.25}\smash{\begin{tabular}[t]{l}Partly (95\%)\end{tabular}}}}%
    \put(0.61678328,0.4465445){\color[rgb]{0,0,0}\makebox(0,0)[lt]{\lineheight{1.25}\smash{\begin{tabular}[t]{l}Closed (87\%)\end{tabular}}}}%
    \put(0.76285286,0.44434263){\color[rgb]{0,0,0}\makebox(0,0)[lt]{\lineheight{1.25}\smash{\begin{tabular}[t]{l}Closed (95\%)\end{tabular}}}}%
    \put(0,0){\includegraphics[width=\unitlength,page=7]{exp_prims_comb_v2.pdf}}%
  \end{picture}%
\endgroup%

%% file: sections/table_2_comb.tex
\begin{table}[!b]
\vspace{-2mm}
    \caption{Quantitative results for unpacked gowns
    using combinations of the three learned primitives and a quasi-static motion. The results indicate the percentage of gowns classified as opened or partly opened according to the trained classifier, and if the arms are positioned Forward (Fwd.). The results also indicate the mean number of iterations (It.) that it takes to reach either the open or partly opened state.} 
    \resizebox{\linewidth}{!}{%
        {\renewcommand\arraystretch{1.4} 
            \centering
            \begin{tabular}{m{0.1\linewidth}|m{0.1\linewidth}|m{0.1\linewidth}|m{0.1\linewidth}|m{0.1\linewidth}}
                % \Cline{1pt}{3-6}
                \multicolumn{1}{c}{ }
                & \multicolumn{1}{|c|}{\parbox[t]{9mm}{\centering Opened}} 
                & \multicolumn{1}{c|}{\parbox[t]{9mm}{\centering Partly Opened}}
                & \multicolumn{1}{c|}{\parbox[t]{9mm}{\centering Arms Fwd.}}
                & \multicolumn{1}{c}{\parbox[t]{5mm}{\centering \# It.}}\\
                \noalign{\hrule height 1pt}
                % Prev Opened
                \multicolumn{1}{c|}{Fling + Shake} 
                & \multicolumn{1}{c|}{0\%} & \multicolumn{1}{c|}{100\%} 
                & \multicolumn{1}{c|}{100\%}
                & \multicolumn{1}{c}{2} \\ \cline{1-5}
                % Row 2
                \multicolumn{1}{c|}{Fling + Twist} 
                & \multicolumn{1}{c|}{0\%} & \multicolumn{1}{c|}{100\%} 
                & \multicolumn{1}{c|}{100\%}
                & \multicolumn{1}{c}{2}  \\ \cline{1-5}
                % Row 3
                \multicolumn{1}{c|}{Twist + Fling} 
                & \multicolumn{1}{c|}{0\%} & \multicolumn{1}{c|}{100\%}
                & \multicolumn{1}{c|}{100\%}
                & \multicolumn{1}{c}{1}  \\ \cline{1-5}
                % Row 4
                \multicolumn{1}{c|}{Twist + Shake} 
                & \multicolumn{1}{c|}{0\%} & \multicolumn{1}{c|}{100\%}
                & \multicolumn{1}{c|}{100\%}
                & \multicolumn{1}{c}{1}  \\ 
                \noalign{\hrule height 1pt}
                % Packed Aprons
                \multicolumn{1}{c|}{Fling + Quasi} 
                & \multicolumn{1}{c|}{0\%} & \multicolumn{1}{c|}{100\%} 
                & \multicolumn{1}{c|}{100\%}
                & \multicolumn{1}{c}{1.5} \\ \cline{1-5}
                % Row 2
                \multicolumn{1}{c|}{Shake + Quasi} 
                & \multicolumn{1}{c|}{0\%} & \multicolumn{1}{c|}{100\%}
                & \multicolumn{1}{c|}{100\%}
                & \multicolumn{1}{c}{1}  \\ \cline{1-5}
                % Row 3
                \multicolumn{1}{c|}{Twist + Quasi}
                & \multicolumn{1}{c|}{0\%} & \multicolumn{1}{c|}{100\%}
                & \multicolumn{1}{c|}{100\%}
                & \multicolumn{1}{c}{1}  \\
                    \noalign{\hrule height 1pt}
            \end{tabular}
        }}
    \label{tab:results-prims-combined}
\end{table}

%% file: sections/discussion.tex
Our results highlight the challenges of unfolding packed medical gowns, where electrostatic forces and tightly folded layers make it difficult to fully open the garment.
While the dynamic motions helped to separate some layers, none of them were sufficient to achieve a fully unfolded state.
Although in theory higher-velocity motions could overcome the electrostatic forces, these high-speed motions would be undesirable in healthcare environments, where safety in the human interactions plays a pivotal role.
Nevertheless, the learned primitives are able to succeed in the opening of previously opened gowns. This suggests that these motions would suffice for unfolding other types of garments, such as t-shirts or trousers, which lack long sleeves that may become tangled when folded.

Our experiments combining multiple motions show that this combination is crucial for achieving a partly opened state. In certain settings, such as assisting nurses in dressing, the partly opened state can be  enough for facilitating gown placement and reducing the nurses burden in repetitive tasks.
Alternative strategies such as air-based manipulation~\cite{xu2022dextairity} could separate the fabric layers by using airflow actions.
However, this type of actions introduce a significant problem in healthcare settings, since airflow actions could interfere with sterile conditions.
For this reason, although the combination of lower-acceleration motions such as the twist with quasi-static motions are not sufficient for reaching the gown opened configuration, they are an effective and safe approach for healthcare environments.